# PALI: A Language Identification Benchmark for Perso-Arabic Scripts

**Sina Ahmadi**     **Milind Agarwal**     **Antonios Anastasopoulos**
Department of Computer Science
George Mason University
{sahmad46,magarwa,antonis}@gmu.edu

## Abstract

The Perso-Arabic scripts are a family of scripts that are widely adopted and used by various linguistic communities around the globe. Identifying various languages using such scripts is crucial to language technologies and challenging in low-resource setups. As such, this paper sheds light on the challenges of detecting languages using Perso-Arabic scripts, especially in bilingual communities where "unconventional" writing is practiced. To address this, we use a set of supervised techniques to classify sentences into their languages. Building on these, we also propose a hierarchical model that targets clusters of languages that are more often confused by the classifiers. Our experiment results indicate the effectiveness of our solutions.[1]

## 1 Introduction

Historically, the territorial expansion of the Arab conquests led to various long-lasting changes in the world, particularly from an ethnolinguistic point of view where the local languages of the time faced existential challenges (Wasserstein, 2003). With Arabic being the language of administration –a *Reichssprache*– many languages were affected and adapted in many ways such as writing or vocabulary. Over centuries the Persian language extended the Arabic script by adding additional graphemes such as <پ> (<p>, U+067E) and <گ> (<g>, U+06AF) to conform to the phonology of the language. Hence, one of the main extended variants of the Classical Arabic script is the Perso-Arabic script which has been gradually adopted by many other languages to our day, mainly in West, Central and South Asia (Khansir and Mozafari, 2014). Some of the languages using a Perso-Arabic script are Urdu, Kurdish, Pashto, Azeri Turkish, Sindhi, and Uyghur, along with many others that historically used the script such as Ottoman

Turkish. This said, there are other scripts that were directly adopted from the Arabic script without being affected by the Persian modifications such as Ajami script used in some African languages like Swahili and Wolof, Pegon and Jawi scripts used in Southern Asia and Aljamiado historically used for some European languages.

Language identification is the task of detecting the language of a text at various levels such as document, sentence and sub-sentence. Given the importance of this task in natural language processing (NLP) as in machine translation and information retrieval, it has been extensively studied and is shown to be beneficial to various applications such as sentiment analysis and machine translation (Jauhiainen et al., 2019). This task is not equally challenging for all setups and languages, as it has been demonstrated that language identification for shorter texts or languages that are closely related, both linguistically and in writing, is more challenging, e.g. Farsi vs. Dari or varieties of Kurdish (Malmasi et al., 2015; Zampieri et al., 2020).

Furthermore, some of the less-resourced languages spoken in bilingual communities face various challenges in writing due to a lack of administrative or educational support for their native language or limited technological tools. These result in textual content written unconventionally, i.e. not according to the conventional script or orthography of the language but relying on that of the administratively "dominant" language. For instance, Kashmiri or Kurdish are sometimes written in the Urdu or Persian scripts, respectively, rather than using their adopted Perso-Arabic orthography. This further complicates the identification of those languages, causing confusion due to the resemblance of scripts and hampers data-driven approaches due to the paucity of data. Therefore, reliable language identification of languages using Perso-Arabic scripts remains a challenge to this day, particularly in under-represented languages.

---

[1] Data and models are available at https://github.com/sinaahmadi/PersoArabicLID

| Language | 639-3 | WP | Script type | Diacritics | ZWNJ | Dominant |
|---|---|---|---|---|---|---|
| Azeri Turkish | `azb` | `azb` | Abjad | ✓ | ✓ | Persian |
| Gilaki | `glk` | `glk` | Abjad | ✓ | ✓ | Persian |
| Mazanderani | `mzn` | `mzn` | Abjad | ✓ | ✓ | Persian |
| Pashto | `pus` | `ps` | Abjad | ✓ | ✗ | Persian |
| Gorani | `hac` | – | Alphabet | ✗ | ✗ | Persian, Arabic, Sorani |
| Northern Kurdish (Kurmanji) | `kmr` | – | Alphabet | ✗ | ✗ | Persian, Arabic |
| Central Kurdish (Sorani) | `ckb` | `ckb` | Alphabet | ✗ | ✗ | Persian, Arabic |
| Southern Kurdish | `sdh` | – | Alphabet | ✗ | ✗ | Persian, Arabic |
| Balochi | `bal` | – | Abjad | ✓ | ✗ | Persian, Urdu |
| Brahui | `brh` | – | Abjad | ✓ | ✗ | Urdu |
| Kashmiri | `kas` | `ks` | Alphabet | ✓ | ✗ | Urdu |
| Sindhi | `snd` | `sd` | Abjad | ✓ | ✗ | Urdu |
| Saraiki | `skr` | `skr` | Abjad | ✓ | ✗ | Urdu |
| Torwali | `trw` | – | Abjad | ✓ | ✗ | Urdu |
| Punjabi | `pnb` | `pnb` | Abjad | ✓ | ✗ | Urdu |
| Persian | `fas` | `fa` | Abjad | ✓ | ✓ | - |
| Arabic | `arb` | `ar` | Abjad | ✓ | ✗ | - |
| Urdu | `urd` | `ur` | Abjad | ✓ | ✓ | - |
| Uyghur | `uig` | `ug` | Alphabet | ✗ | ✗ | - |

Table 1: Perso-Arabic scripts of the selected languages studied in this paper. Columns 2 and 3 show the codes of the languages in ISO 639-3 and on their specific Wikipedia (WP), if available. The diacritics and zero-width non-joiner (ZWNJ) columns refer to the usage of diacritics (*Harakat*) and ZWNJ as individual characters.

As such, we select several languages that use Perso-Arabic scripts, summarized in Table 1. Among these, the majority face challenges related not only to a scarcity of data but also unconventional writing. Therefore, we define the language identification task for these languages in two setups where a) the text is written according to the script or orthography of the language, referred to as conventional writing, or b) the text contains a certain degree of anomalies due to usage of the script or orthography of the administratively-dominant language. Considering that Perso-Arabic scripts are mostly used in languages native to Pakistan, Iran, Afghanistan and Iraq, we also include Urdu, Persian and Arabic as they are primarily used as administratively-dominant languages. Furthermore, having a more diverse set of languages can reveal which languages are more often confused. Although we also include Uyghur, it should be noted that it is mainly spoken in a bilingual community, i.e. in China, where unconventional writing is not Perso-Arabic; therefore, we only consider conventional writing for Uyghur.

**Contributions** This paper sheds light on language identification for languages written in the Perso-Arabic script or its variants. We describe collecting data and generating synthetically-noisy sentences using script mapping (§2). We implement a few classification techniques and propose a hierarchical model approach to resolve confusion between clusters of languages. The proposed approach outperforms other techniques with a macro-average $F_1$ that ranges from 0.88 to 0.95 for noisy settings (§3).

## 2 Methodology

Given that the selected languages are mostly low-resourced, collecting data and, more importantly, identifying text written in a conventional and unconventional way is a formidable task. To tackle this, we focus on collecting data from various sources on the Web, notably Wikipedia.[2] Then, we propose an approach to generate synthetic data that can potentially reflect and model various types of noise that occur in unconventional writing. To that end, we use a simple technique that maps characters from the script of a language to that of another one, i.e. the dominant language. And finally, we discuss our efforts to benchmark this task and propose a hierarchical model that resolves confusion between similar languages.

---

[2] https://www.wikidata.org

## 2.1 Data Collection

As Table 1 shows, all languages have their dedicated Wikipedia pages using their Perso-Arabic scripts, except Gorani, Northern and Southern Kurdish, Balochi, Brahui and Torwali. Therefore, we use the Wikipedia dumps as corpora for the available languages.[3] On the other hand, for Northern and Southern Kurdish, Balochi and Brahui, we collect data by crawling local news websites as listed in Table A.2. Additionally, we use Uddin and Uddin (2019)'s corpus for Torwali, Ahmadi (2020)'s corpus for Gorani, Esmaili et al. (2013)'s corpus for Central Kurdish Tehseen et al. (2022)'s corpus for Punjabi. Regarding Persian, Arabic and Urdu, we use the Tatoeba datasets.[4]

Once the data is collected, we carry out text preprocessing after converting various formats to raw text, use regular expressions to remove special characters related to formatting styles and remove information such as emails, phone numbers, and website URLs. We also convert numerals to Latin ones as a mixture of numerals is usually used in Perso-Arabic texts, namely Persian <۰۱۲۳۴۵۶۷۸۹> and Arabic <٠١٢٣٤٥٦٧٨٩> numerals along with the Latin ones. This is to ensure that a diverse set of numerals are later included in the sentences for the language identification task. As some of the selected languages use two scripts, as in Punjabi written in Gurmukhi and Shahmukhi or Kashmiri written in Devanagari and Perso-Arabic, we also applied a few regular expressions to remove script and code-switched sentences or quoted ones in the corpora. Given the complexity of detecting such alternations, we note that script and code-switched words may still exist in the cleaned corpora.

We finalize text preprocessing by unifying the Unicode encoding of characters. Inconsistencies in Unicode encoding are oftentimes due to the usage of keyboards with different code bindings and are previously included in preprocessing for some languages (Ahmadi, 2019; Doctor et al., 2022). As an example, <ے> (U+06D2) and <ي> (U+064A) may be used instead of <ی> (U+06CC) or <ک> (U+0643) instead of <ک> (U+06A9) in Kurdish. Depending on the usage of zero-width non-joiner character (ZWNJ, U+200C), as shown in Table 1, we also consider its removal in the preprocessing step.[5] Finally, we tokenize the corpora at the sentence level using regular expressions.

Table A.3 presents the 10 most frequent trigrams in the collected corpora among which many affixes and conjunctions are retrieved that can be indicative of a language.

## 2.2 Script Mapping

Assuming that a noisy text is written using the dominant language's script or orthography, we map the Perso-Arabic script of a given language to that of the dominant language, e.g. Kashmiri script to Urdu or Central Kurdish script to Persian and Arabic. To do so, we rely on the visual resemblance and Unicode encoding of the characters as follows:

- If two graphemes exist in the scripts of the two languages, as in <ھ> (U+06BE) in Sindhi and Urdu or <ٹ> (U+0679) in Saraiki and Urdu, we map them together regardless of their pronunciation in the two languages.

- In absence of an identical grapheme in the dominant script, the most visually similar character is mapped to the source character. For instance, the most similar character in Urdu to <ڷ> (U+06B7) in Brahui is <ل> (U+0644). Similarly, <ۇ> (U+06CB) in Gilaki is mapped to the similar <و> (U+0648) in Persian. This way, a character can be mapped to many other characters in the source language.

- Some mappings follow orthographic rules, particularly for characters that vary depending on the position in a word. For instance, vowels in Kurdish appear with an initial *hamza*, i.e. <ئ> (U+0626) as in <ئۆ> /oː/ and <ئێ> /eː/. We also include such rules.

- Since the numerals are unified in data collection (§2.1), we also map the Latin numerals to those of Persian and Arabic randomly.

Depending on the dominant languages, for each source and dominant language, a script mapping is manually created. It should be noted that along with the non-diacritical characters, diacritical ones are also included if the diacritics, including *Harakat*, are part of the grapheme as in <ڎ> (U+068E) in Gorani and Sindhi, but not <ڌ>. Detachable Harakat such as *fatha*, *kasra* and *damma* are not included in the script mapping. Table A.1 presents the set of characters used in the selected languages based on their relation with Arabic, Persian, and Urdu as the three major languages using Perso-Arabic scripts.

---

[3] Dumps of 20 January 2023.

[4] https://tatoeba.org

[5] We consult various sources on the Web for information about common writing practices in the selected languages, notably https://scriptsource.org.

### 2.3 Synthetic Data Generation

Using the script mappings, we mimic unconventional writing by generating synthetic sentences based on the 'clean' ones, i.e. sentences in the collected corpora. This is carried out by randomly substituting characters in the clean sentence with an alternative in the target script using our mappings. In order to evaluate the impact of noise on language identification, we synthesize data at various levels starting from 20% noise up to 100%, where a certain level of noise is applied based on the number of possible substitutions. Table 2 shows an example of a clean sentence in Northern Kurdish and its synthetic noisy equivalents based on the level of noise.

Therefore, the datasets are categorized as follows:

1. CLEAN: a dataset containing original sentences from the corpora without injecting any noise. This is equivalent to 0% of noise in the data. This includes all the selected languages along with Urdu, Persian, Arabic, and Uyghur.

2. NOISY: datasets of sentences having noisy characters at various levels, starting from 20% of noise and gradually increasing 20% up to 100%. Regardless of usage, detachable diacritics are removed when the noise level is 100%, including for Kashmiri for which diacritics are strictly used. We combine all data with all levels of noise in a separate dataset called ALL. Given that Persian, Urdu, Arabic, and Uyghur do not face unconventional writing, they are not included in the noisy data.

3. MERGED: the result of merging CLEAN and ALL datasets.

The CLEAN and NOISY datasets contain 10,000 sentences per language, except for Brahui, Torwali, and Balochi, for which only 549, 1371, and 1649 sentences are available in the corpora respectively. Therefore, we included 500 sentences from those languages in the test sets and upsample the remaining sentences with a coefficient of four, i.e. duplicating four times the remaining sentences, and consider them as a train set. Similarly, for Kashmiri and Gorani for which 6340 and 8742 sentences are respectively available, 2000 sentences are added to the test set while the remaining sentences are upsampled to have 8000 sentences in the train set.

To avoid an imbalance of data for dominant languages for which there is no noise, i.e. Urdu, Per-sian, Arabic, along with Uyghur, 10,000 more instances are added from their respective clean corpora. As such, the MERGED dataset contains 20,000 clean and noisy sentences per language.

| Noise % | Sentence |
|---------|----------|
| Clean | دووەمین پێشانگەها فۆتۆگرافەرێن کورد ل بەلجیکا |
| | Second Kurdish photographers' exhibition in Belgium |
| 20 | دووەمین پێشانگەها فۆتۆگرافەرێن کورد ل بەلجیکا |
| 40 | دووە مین پشانگە ها فطکرافە رن کورد ل بە لجیکا |
| 60 | دووە مین پشانگە ها فۆتۆگرافە رن کورد ل بە لجیکا |
| 80 | دووەمین پیشانگەها فۆتۆگرافەرین کورد ل بەلجیکا |
| 100 | دووەمین پیشانگەها فۆتۆگرافەرین کورد ل بەلجیکا |

Table 2: A sentence in Northern Kurdish (Kurmanji) along with its synthetically-generated noisy ones based on different levels of noise.

### 2.4 Benchmarking

We consider language identification as a probabilistic classification problem where each sentence is predicted to belong to a specific class, i.e. language, with a certain probability. We use the 80/20 split of the sentences in the various datasets for the train and test sets as described in the previous sections. Both sets are from the same data.

As a baseline system, we use fastText's pretrained language identification model–lid.176 that is trained using data from Wikipedia, Tatoeba and SETimes for 176 languages, including all the selected languages except Balochi, Brahui, Gilaki, Gorani, Northern Kurdish (in Perso-Arabic script), Southern Kurdish and Torwali. In addition, we train a model using fastText with word vectors of size 64, a minimum and maximum length of character $n$-grams of 2 to 6, 1.0 learning rate, 25 epochs and a hierarchical softmax loss.

Other than the fastText-related baseline and our own models, we also report precision, recall, and $F_1$ scores for benchmarking purposes for state-of-the-art methods such as Google's CLD3 (Salcianu et al., 2020), Franc[6] and Langid.py (Lui and Baldwin, 2012). We also share two other baselines trained from scratch with character $n$-gram features of sizes 2 to 4 - Multinomial Naive Bayes model (MNB – non-uniform learned class priors, no Laplace smoothing), and a Multilayer Perceptron (MLP) with maximum iterations of 500, one hidden layer of size 500 and a batch size of 1000.

---

[6] https://github.com/wooorm/franc/

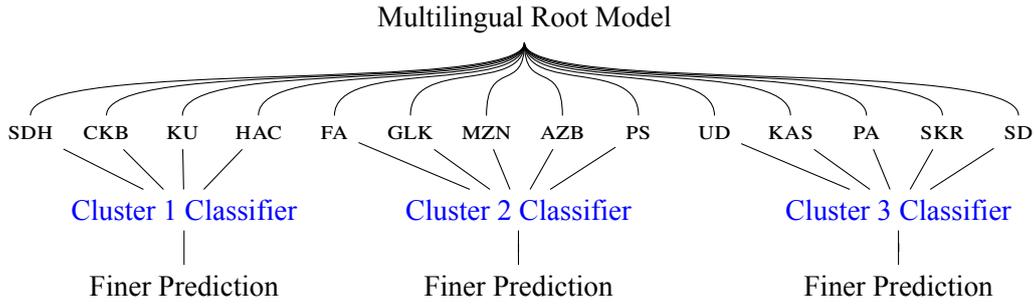

Figure 1: Architecture of our hierarchical model. If the root model predicts Southern Kurdish (ꜱᴅʜ), Gorani (ʜᴀᴄ), Northern Kurdish (ᴋᴍʀ), or Central Kurdish (ᴄᴋʙ), the sample gets sent down to a smaller expert classifier that is trained to resolve confusion between these four closely-related languages. Likewise for cluster 2 and cluster 3's languages. If an unclustered language is predicted by the root model, i.e. none of the branches are available, the hierarchical model predicts the same label as the root.

| Prediction \ True | Southern Kurdish | Central Kurdish | Northern Kurdish | Gorani | Persian | Gilaki | Mazanderani | Azeri Turkish | Pashto | Urdu | Kashmiri | Punjabi | Sindhi | Saraiki | Arabic | Balochi | Torwali | Uyghur | Brahui |
|---|---|---|---|---|---|---|---|---|---|---|---|---|---|---|---|---|---|---|---|
| Southern Kurdish | 15643 | 99 | 70 | 113 | 0 | 2 | 0 | 0 | 1 | 0 | 0 | 0 | 0 | 0 | 2 | 0 | 0 | 0 | 0 |
| Central Kurdish | 242 | 15850 | 94 | 64 | 0 | 1 | 1 | 2 | 0 | 0 | 0 | 1 | 1 | 0 | 0 | 0 | 0 | 1 | 0 |
| Northern Kurdish | 49 | 29 | 15800 | 41 | 1 | 0 | 0 | 6 | 2 | 0 | 0 | 1 | 2 | 0 | 1 | 0 | 0 | 5 | 0 |
| Gorani | 59 | 21 | 18 | 15746 | 0 | 3 | 4 | 3 | 0 | 0 | 0 | 1 | 1 | 1 | 1 | 0 | 0 | 6 | 0 |
| Persian | 2 | 0 | 0 | 2 | 15874 | 50 | 26 | 7 | 8 | 0 | 3 | 1 | 2 | 2 | 7 | 0 | 0 | 0 | 0 |
| Gilaki | 2 | 0 | 2 | 10 | 63 | 15778 | 129 | 66 | 1 | 0 | 3 | 1 | 18 | 1 | 3 | 1 | 0 | 1 | 0 |
| Mazanderani | 0 | 0 | 0 | 3 | 18 | 92 | 15709 | 72 | 7 | 0 | 7 | 2 | 3 | 2 | 4 | 0 | 0 | 1 | 0 |
| Azeri Turkish | 0 | 0 | 2 | 6 | 1 | 44 | 91 | 15772 | 22 | 4 | 4 | 11 | 4 | 0 | 1 | 0 | 1 | 1 | 0 |
| Pashto | 2 | 1 | 7 | 3 | 21 | 2 | 6 | 34 | 15916 | 1 | 7 | 14 | 16 | 3 | 1 | 3 | 1 | 3 | 1 |
| Urdu | 0 | 0 | 0 | 0 | 0 | 0 | 0 | 0 | 0 | 15902 | 4 | 78 | 24 | 32 | 0 | 2 | 14 | 0 | 1 |
| Kashmiri | 0 | 0 | 1 | 0 | 0 | 3 | 8 | 7 | 7 | 3 | 15889 | 28 | 17 | 21 | 2 | 0 | 0 | 2 | 0 |
| Punjabi | 0 | 0 | 0 | 0 | 2 | 1 | 5 | 8 | 14 | 33 | 33 | 15782 | 26 | 95 | 0 | 7 | 8 | 1 | 0 |
| Sindhi | 0 | 0 | 1 | 2 | 1 | 16 | 5 | 1 | 0 | 10 | 5 | 12 | 15800 | 13 | 17 | 1 | 0 | 0 | 0 |
| Saraiki | 0 | 0 | 0 | 1 | 8 | 1 | 5 | 11 | 4 | 32 | 37 | 62 | 34 | 15818 | 0 | 14 | 6 | 2 | 0 |
| Arabic | 1 | 0 | 1 | 1 | 10 | 5 | 7 | 8 | 9 | 0 | 8 | 0 | 43 | 1 | 15955 | 1 | 0 | 12 | 0 |
| Balochi | 0 | 0 | 0 | 0 | 0 | 1 | 1 | 0 | 1 | 1 | 0 | 0 | 6 | 1 | 1 | 7464 | 0 | 0 | 0 |
| Torwali | 0 | 0 | 0 | 0 | 0 | 1 | 0 | 0 | 12 | 0 | 4 | 2 | 8 | 0 | 0 | 0 | 3590 | 0 | 0 |
| Uyghur | 0 | 0 | 4 | 8 | 0 | 0 | 3 | 3 | 1 | 1 | 0 | 0 | 0 | 0 | 5 | 0 | 0 | 15965 | 0 |
| Brahui | 0 | 0 | 0 | 0 | 0 | 0 | 0 | 0 | 0 | 3 | 1 | 2 | 0 | 0 | 0 | 0 | 0 | 0 | 286 |

Figure 2: Confusion matrix of the multilingual root model on the training dataset. Row labels indicate our custom fastText model's predictions, columns indicate true labels (training dataset), and each cell count indicates the number of predictions made by the model for a (prediction, true label) pair. From the confusion matrix, we identified three highly-confused language clusters as reported in Section 2.5.

### 2.5 Hierarchical Modeling

The goal behind hierarchical modeling (Figure 1) is to resolve a model's confusion between highly-related languages by training expert classifiers that specialize in distinguishing between a small set of languages. We achieve this by inspecting the confusion matrix of the best-performing model (on training data) and identifying language clusters that the model shows high confusion in predicting. The custom-trained fastText model described in the previous section serves as the root classifier and we identify three clusters, as mentioned below, from its confusion matrix (Figure 2):

1. Cluster 1 containing Southern Kurdish, Central Kurdish, Northern Kurdish and Gorani
2. Cluster 2 containing Persian, Gilaki, Mazanderani, Azeri Turkish and Pashto
3. Cluster 3 containing Urdu, Kashmiri, Punjabi, Sindhi and Saraiki

Each sub-unit in the hierarchical tree is a fast-Text model trained from scratch on data from the relevant cluster's languages with the same parameterization as the root model.

### 3 Results

In Table 3, we report precision, recall, and $F_1$ scores across all datasets, 6 state-of-the-art and custom-trained baselines, our root fastText model (Root), and a hierarchical confusion-resolution model (Hier). We find that our root fastText model performs well by considerable margins when compared to the pre-trained fastText baseline, Google's CLD3, `langid.py`, `Franc`, MNB and MLP.

#### 3.1 State-of-the-Art vs. Simple Baselines

None of the three state-of-the-art models (CLD3, `langid.py`, `Franc`) get more than 0.15 $F_1$ score on our test set across all 19 languages and noise settings. In fact, they often get acutely low $F_1$ scores ($0 \leq F_1 < 0.1$) for mixed noise settings (40% - ALL). This is despite these models' support of Urdu, Persian, Arabic, Sindhi, with `Franc` additionally covering Central Kurdish. This demonstrates the poor quality of language identification in the state-of-the-art pre-trained models despite claims of covering hundreds of languages, further highlighting that language identification is far from solved. Compared to these three models, the MNB and MLP models perform better across all noise levels (except 20% noise), and even outperform fastText's large pre-trained model `lid.176` on 7 out of 8 noise settings, becoming a stronger baseline than the `lid.176` model.

#### 3.2 Hierarchical Modeling with fastText

Coming to our two models, the custom fast-Text model (Root) and the hierarchical confusion-resolution model (Hier), it is clear that both models perform noticeably better compared to any of the baselines by a huge margin. Since the hierarchical model is trained on the MERGED dataset which contains noisy and clean sentences with four more classes than the clean (0% noise) setting, it is natural that the Root model performs better in the clean setting. However, for any realistic noise level (from 20% to MERGED) the hierarchical model performs better than the Root model.

To test these subtle improvements, we report statistical significance results for each noise level according to a one-tailed Z-test, comparing the root model with the hierarchical model, at a significance level 0.01. We perform a Z-test because the number of samples is greater than 30 and the sample variance can be reliably used as an estimate of the population variance. The null hypothesis is that there is no significant difference between the root and the hierarchical model ($\mu_0 : f_{root} = f_{hier}$) and the alternative hypothesis proposes that the hierarchical model's performance is significantly and strictly greater than the root model ($\mu_1 : f_{root} < f_{hier}$). We compute a one-tailed 99% confidence interval for the root model's $F_1$ score $f_{root}$. As per the one-tailed Z-test, we can reject the null hypothesis and conclude that the difference between $F_1$ scores is statistically significant if the hierarchical model's $F_1$ score $f_{hier}$ is strictly over this interval's upper bound. In Table 4, we report the results of our hypothesis testing and find that the advantage provided by our hierarchical confusion-resolution approach is statistically significant at the 99% confidence level for all noise settings. Therefore, we establish that a confusion-informed hierarchical approach could be utilized to improve performance on noisy data without re-training the entire model and that it translates well to the test set by bringing statistically significant improvements.

#### 3.3 Language-Specific Performance

In Table 5, we report language-level scores across noise levels for the best two systems: our custom fastText model and our confusion-resolution hierarchical model. Across all languages and noise levels, the hierarchical model only underperforms in

| Noise | Metric | Hier | Root | fastText | CLD3 | langid.py | Franc | MNB | MLP |
|---|---|---|---|---|---|---|---|---|---|
| 0% | Precision | 0.72 | **0.91** | 0.16 | 0.03 | 0 | 0.02 | 0.43 | 0.47 |
| | Recall | 0.70 | **0.89** | 0.07 | 0.05 | 0 | 0.02 | 0.14 | 0.16 |
| | F$_1$ Score | 0.72 | **0.90** | 0.10 | 0.04 | 0 | 0.02 | 0.21 | 0.24 |
| 20% | Precision | **0.92** | 0.92 | 0.30 | 0.08 | 0.13 | 0.13 | 0.08 | 0.03 |
| | Recall | **0.89** | 0.89 | 0.32 | 0.18 | 0.18 | 0.18 | 0.05 | 0.05 |
| | F$_1$ Score | **0.91** | 0.90 | 0.31 | 0.11 | 0.15 | 0.15 | 0.06 | 0.04 |
| 40% | Precision | **0.91** | 0.90 | 0.17 | 0.04 | 0 | 0.01 | 0.51 | 0.49 |
| | Recall | **0.88** | 0.88 | 0.07 | 0.05 | 0 | 0 | 0.09 | 0.11 |
| | F$_1$ Score | **0.90** | 0.89 | 0.10 | 0.05 | 0 | 0 | 0.16 | 0.19 |
| 60% | Precision | **0.91** | 0.90 | 0.17 | 0.04 | 0 | 0 | 0.45 | 0.54 |
| | Recall | **0.88** | 0.87 | 0.07 | 0.05 | 0 | 0 | 0.12 | 0.09 |
| | F$_1$ Score | **0.89** | 0.88 | 0.09 | 0.04 | 0 | 0 | 0.20 | 0.15 |
| 80% | Precision | **0.90** | 0.90 | 0.16 | 0.03 | 0 | 0 | 0.25 | 0.33 |
| | Recall | **0.88** | 0.87 | 0.06 | 0.05 | 0 | 0 | 0.12 | 0.15 |
| | F$_1$ Score | **0.89** | 0.88 | 0.08 | 0.04 | 0 | 0 | 0.16 | 0.21 |
| 100% | Precision | **0.90** | 0.90 | 0.15 | 0.03 | 0 | 0 | 0.44 | 0.44 |
| | Recall | **0.88** | 0.87 | 0.06 | 0.05 | 0 | 0 | 0.08 | 0.11 |
| | F$_1$ Score | **0.89** | 0.88 | 0.08 | 0.03 | 0 | 0 | 0.13 | 0.17 |
| ALL | Precision | **0.90** | 0.89 | 0.15 | 0.03 | 0 | 0 | 0.28 | 0.51 |
| | Recall | **0.87** | 0.86 | 0.06 | 0.05 | 0 | 0 | 0.16 | 0.10 |
| | F$_1$ Score | **0.88** | 0.88 | 0.08 | 0.04 | 0 | 0 | 0.20 | 0.17 |
| MERGED | Precision | **0.95** | 0.95 | 0.28 | 0.06 | 0.11 | 0.11 | 0.15 | 0.15 |
| | Recall | **0.94** | 0.94 | 0.27 | 0.16 | 0.16 | 0.16 | 0.08 | 0.07 |
| | F$_1$ Score | **0.95** | 0.94 | 0.27 | 0.09 | 0.13 | 0.13 | 0.10 | 0.10 |

Table 3: Comparison of all language identification models' precision, recall, and F$_1$ scores across noise settings. Our hierarchical (Hier) and Root models perform as the best two models for all noise levels. fastText, Multinomial Naive Bayes (MNB) and Multilayer Perceptron (MLP) take third place for different noise levels. Precision, recall, and F$_1$ scores are reported for all methods to provide benchmarks. For two values that are the same up to the hundredth decimal place, boldfaced entries indicate strictly better performance.

| Noise | Test Samples | $\Delta$ | Significant |
|---|---|---|---|
| 0 | 33500 | -0.188 | ✗ |
| 20 | 25500 | 0.005 | ✓ |
| 40 | 25500 | 0.006 | ✓ |
| 60 | 25500 | 0.007 | ✓ |
| 80 | 25500 | 0.007 | ✓ |
| 100 | 25500 | 0.007 | ✓ |
| ALL | 27806 | 0.007 | ✓ |
| MERGED | 69304 | 0.002 | ✓ |

Table 4: Improvements (positive $\Delta$) in the F$_1$ scores of our hierarchical modeling approach compared to the Root model are statistically significant for all noise levels at significance level = 0.01, i.e. 99% confidence.

5 out of 128 settings. For all others, it performs either at par or better than the Root model. The boldface entries indicate that the hierarchical model brings the most improvements in the noisy settings (20%-ALL) across all three identified clusters. As expected, for languages that were not part of any highly-confused cluster, i.e. AR, BAL, TRW, UG and BRH, the hierarchical and Root model produce the same predictions, therefore, have the same scores across noise levels. In Table A.4, we also provide a few language identification examples at various noise levels based on the predictions of the pre-trained fastText model in comparison to our model.

| | 0% | | 20% | | 40% | | 60% | | 80% | | 100% | | ALL | | MERGED | |
|---|---|---|---|---|---|---|---|---|---|---|---|---|---|---|---|---|
| | M1 | M2 | M1 | M2 | M1 | M2 | M1 | M2 | M1 | M2 | M1 | M2 | M1 | M2 | M1 | M2 |
| **Cluster 1** | | | | | | | | | | | | | | | | |
| SDH | 0.95 | **0.96** | 0.95 | **0.96** | 0.94 | **0.95** | 0.93 | **0.94** | 0.93 | **0.94** | 0.94 | 0.94 | 0.94 | 0.94 | 0.95 | **0.96** |
| CKB | 0.95 | 0.95 | 0.94 | 0.94 | 0.92 | **0.94** | 0.91 | **0.93** | 0.91 | **0.93** | 0.91 | **0.92** | 0.92 | **0.93** | 0.95 | 0.95 |
| KU | 0.95 | 0.95 | 0.93 | **0.94** | 0.93 | 0.93 | 0.92 | **0.93** | 0.93 | 0.92 | 0.92 | 0.92 | 0.92 | **0.93** | 0.95 | 0.95 |
| HAC | 0.94 | 0.94 | 0.94 | 0.94 | 0.93 | 0.93 | 0.92 | 0.92 | 0.92 | 0.92 | 0.92 | 0.92 | 0.91 | **0.92** | 0.94 | 0.94 |
| **Cluster 2** | | | | | | | | | | | | | | | | |
| FA | 0.97 | **0.98** | - | - | - | - | - | - | - | - | - | - | - | - | 0.97 | **0.98** |
| GLK | 0.92 | **0.94** | 0.88 | **0.89** | 0.88 | **0.89** | 0.88 | **0.9** | 0.88 | **0.89** | 0.88 | **0.89** | 0.92 | 0.92 | 0.92 | **0.94** |
| MZN | 0.92 | 0.92 | 0.85 | **0.86** | 0.85 | **0.86** | 0.85 | **0.87** | 0.85 | **0.86** | 0.85 | **0.87** | 0.92 | **0.93** | 0.92 | 0.92 |
| AZB | 0.91 | 0.91 | 0.86 | **0.87** | 0.85 | **0.86** | 0.86 | **0.87** | 0.86 | **0.87** | 0.85 | **0.86** | 0.9 | **0.91** | 0.91 | 0.91 |
| PS | 0.96 | 0.96 | 0.94 | **0.95** | 0.94 | **0.95** | 0.94 | **0.95** | 0.94 | **0.95** | 0.94 | 0.94 | 0.96 | 0.96 | 0.96 | 0.96 |
| **Cluster 3** | | | | | | | | | | | | | | | | |
| UD | 0.96 | **0.97** | - | - | - | - | - | - | - | - | - | - | - | - | 0.96 | **0.97** |
| KAS | 0.94 | **0.95** | 0.9 | **0.91** | 0.9 | **0.91** | 0.9 | **0.91** | 0.9 | **0.91** | 0.87 | **0.88** | 0.91 | 0.9 | 0.94 | **0.95** |
| PA | 0.91 | 0.91 | **0.87** | 0.86 | 0.86 | 0.86 | 0.86 | 0.85 | 0.85 | **0.86** | 0.85 | 0.85 | 0.87 | 0.87 | 0.91 | 0.91 |
| SD | 0.93 | **0.94** | 0.89 | **0.91** | 0.88 | **0.89** | 0.87 | **0.89** | 0.87 | **0.89** | 0.87 | **0.89** | 0.91 | 0.91 | 0.93 | **0.94** |
| SKR | **0.92** | 0.91 | 0.85 | 0.85 | 0.84 | **0.85** | 0.84 | **0.85** | 0.85 | 0.85 | 0.84 | **0.85** | 0.86 | **0.88** | **0.92** | 0.91 |
| AR | 0.98 | 0.98 | - | - | - | - | - | - | - | - | - | - | - | - | 0.98 | 0.98 |
| BAL | 0.98 | 0.98 | 0.94 | 0.94 | 0.94 | 0.94 | 0.94 | 0.94 | 0.95 | 0.95 | 0.95 | 0.95 | 0.97 | 0.97 | 0.98 | 0.98 |
| TRW | 0.95 | 0.95 | 0.87 | 0.87 | 0.89 | 0.89 | 0.88 | 0.88 | 0.88 | 0.88 | 0.87 | 0.87 | 0.91 | 0.91 | 0.95 | 0.95 |
| UG | 0.99 | 0.99 | - | - | - | - | - | - | - | - | - | - | - | - | 0.99 | 0.99 |
| BRH | 0.84 | 0.84 | 0.7 | 0.7 | 0.67 | 0.67 | 0.68 | 0.68 | 0.68 | 0.68 | 0.65 | 0.65 | 0.63 | 0.63 | 0.84 | 0.84 |

Table 5: Language-level $F_1$ scores for our hierarchical (M1) and Root (M2) models. Our hierarchical model shows improvement in $F_1$ score for languages in all three clusters (first 3 sections from the top) across noise levels. Dashed cells show that the language only has a conventional script and therefore was not part of the synthetic data settings.

## 4 Related Work

**Modeling Approaches** Language identification is generally modeled as a multi-class text classification task and has achieved state-of-the-art performance with straightforward byte, character or word-level $n$-gram features across languages and language varieties and in limited data settings (Jauhiainen et al., 2017). Model or classifier choice is highly dependent on the source, domain and quantity of data per language, with simple linear classifiers like Support Vector Machines (Ciobanu et al., 2018; Malmasi and Dras, 2015) and Multinomial Naive Bayes (King et al., 2014; Mathur et al., 2017) providing strong baselines with limited data and compute across domains. If large amounts of data are available, aggregated classifiers (Baimukan et al., 2022) and neural models may be used, but have continued to struggle with similar language varieties and dialects and have been prone to overfitting (Medvedeva et al., 2017; Criscuolo and Aluísio, 2017; Eldesouki et al., 2016).

In our paper, we propose a hierarchical approach to language identification that identifies commonly-confusable language pairs in noisy settings and resolves such mispredictions with small classification units. Such a modeling approach can be used to expand language coverage and improve the performance of the existing pre-trained models without retraining large compute-hungry models. In our case, we noticed statistically significant improvements for noisy data settings.

**Similar Languages and Varieties** Language identification is a well-studied problem, sometimes even considered *solved*; in reality, most of the world's languages are not supported by current systems. This lack of representation affects large-scale data mining efforts and further exacerbates data shortage for low-resource languages. One key bottleneck in improving language coverage in language identification systems is the ability to distinguish between similar languages, language varieties, and dialects. As outlined in this paper, this becomes even more challenging when a language community adopts the unconventional script of a dominant language. Recently, there has been studies in distinguishing between Nordic languages (Haas and Derczynski, 2021),

Arabic dialects (Nayel et al., 2021; Abdul-Mageed et al., 2020; Salameh et al., 2018) and regional Italian and French language varieties (Jauhiainen et al., 2022; Camposampiero et al., 2022). Haas and Derczynski (2021), for instance, experiment with many modeling and featurization approaches to best distinguish between six Nordic languages: Danish, Swedish, Norwegian (Nynorsk), Norwegian (Bokmål), Faroese and Icelandic. They find that skipgram embeddings extracted out of fastText are rich and capable of distinguishing between closely-related languages. It is worth noting that while the paper's approach presented improvements across selected languages, all six selected Nordic languages have a large amount of training data (50K+ sentences) and are already supported by off-the-shelf tools like `langid.py`. This is in contrast to our work where previously unsupported languages and varieties are incorporated into language identification systems and evaluated.

To distinguish between similar languages and dialects, more shallow and linear classifiers such as Naive Bayes and Logistic Regression tend to outperform neural models like MLP or convolutional neural networks (Chakravarthi et al., 2021; Aepli et al., 2022; Ceolin, 2021). This is confirmed by non-neural classical machine learning approaches winning a majority of VarDial 2021 and 2022 shared tasks across typologically diverse languages such as Dravidian languages, Romanian dialects, Italian and French regional varieties (Jauhiainen et al., 2022; Camposampiero et al., 2022), and Uralic languages (Chakravarthi et al., 2021). Neural modeling approaches, due to limited data in similar languages/varieties, may also sometimes under-perform non-neural baselines as reported in the Uralic Language Identification or the Italian Dialect Identification shared tasks (Chakravarthi et al., 2021; Aepli et al., 2022).

## 5 Conclusion

We focus our study on languages written in bilingual communities where an unconventional dominant Perso-Arabic script is often utilized in place of a conventional and more suitable Perso-Arabic variant writing system. We discuss challenges unique to this scenario, in both data collection and language identification, and consequent performance issues in state-of-the-art systems when faced with data in such unconventional writing systems. This is highlighted by the 20-point perfor-mance difference in $F_1$ scores between noisy and clean/mixed settings. Our proposed hierarchical approach outperforms a custom-trained fastText system, simple MNB and MLP and the state-of-the-art language identification systems of Google's CLD3, `Franc` and `langid.py`. We find statistically significant improvements by using a hierarchical model after analyzing a root multilingual model's confusion matrix.

## 6 Limitations

Some of the selected languages use more than one script, as in Punjabi or Kurdish. This affects the quality of the collected data which is preprocessed automatically. As such, we believe that our datasets contain a trivial but existing amount of code-switched text. Moreover, having focused on the Perso-Arabic scripts, we did not include texts from other scripts of such languages. Although a language can be affected by more than one dominant language and the synthetic data is generated by considering various script mappings, the impact of individual dominant languages is yet to be analyzed. To this end, a finer-grained classification task should be defined per dominant language.

Additionally, variants such as Dari and Farsi of Persian, and sub-dialects of the selected languages could be included in this task. In the same vein, our hierarchical approach can be applied to other scripts, particularly those that are adopted by many languages, such as Cyrillic and Latin. Finally, other techniques can be implemented and fine-tuned based on our collected data.

Generally, it is expected that the presented models perform better when trained on more data. We also believe that our hierarchical model's improvements over the root model are limited by the size of our training sets. With more genuine noisy data available, it is possible that our performance will improve across all noise setups as well as the clean data setup.

## Acknowledgments

This work was generously supported by the National Science Foundation under DEL/DLI award BCS-2109578, and by the National Endowment for the Humanities under award PR-276810-21. The authors are also grateful to the anonymous reviewers, as well as the Office of Research Computing at GMU, where all computational experiments were conducted.

# A  Selected Languages

| Graphemes | | Kurdish | Gorani | Uyghur | Arabic | Azeri | Gilaki | Mazanderani | Persian | Urdu | Kashmiri | Punjabi | Saraiki | Torwali | Pashto | Sindhi | Brahui | Balochi |
|---|---|---|---|---|---|---|---|---|---|---|---|---|---|---|---|---|---|---|
| Arabic | | | | | | | | | | | | | | | | | | |
| Persian | | | | | | | | | | | | | | | | | | |
| Urdu | | | | | | | | | | | | | | | | | | |
| Language-specific | | | | | | | | | | | | | | | | | | |
| Total/Unique | | 35/7 | 40/12 | 34/7 | 35/0 | 41/4 | 40/3 | 38/1 | 37/0 | 43/0 | 50/6 | 46/3 | 48/5 | 49/8 | 52/13 | 58/21 | 44/1 | 32/3 |

Table A.1: The Perso-Arabic scripts used in the selected languages with a comparative overview of the Arabic, Persian and Urdu scripts. Note that language-specific characters refer to those characters that are unique to a language and not used in Arabic, Persian or Urdu. This is shown in the last row as well.

| Language | Website |
|---|---|
| Balochi | `https://sunnionline.us/balochi/` |
| Brahui | `https://talarbrahui.com` |
| Northern Kurdish | `https://badinan.org` |
| Southern Kurdish | `https://shafaq.com/ku` |

Table A.2: Local news websites from which the collected data are crawled.

| KMR | CKB | SDH | HAC | UIG | ARB | AZB | GLK | MZN | FAS | URD | KAS | PNB | SKR | TRW | PUS | SND | BRH | BAL |
|---|---|---|---|---|---|---|---|---|---|---|---|---|---|---|---|---|---|---|
| ـوبن | لهی | لهب | بهی | نلك | أنی | بير | ئای | بان | ـوین | بین | چهی | یسی | یسی | بنا | بجی | بنا | ئنا | ـبنت |
| بین | لهی | لهب | بهی | ناك | أنی | ستا | بنه | ـوین | است | ـیس | آب | بان | ـبت | بان | بهه | ـبان | نان | نان |
| همی | وهی | انی | شی | بنك | منی | ئان | یتا | بتا | وین | من | فن | ـوین | بهی | ئوب | بان | اسب | بنی | انا |
| ئهی | بهی | چهی | بای | نبی | الم | اوب | هته | ویه | ـبدا | بین | ـبه | بنی | همی | ناو | کین | بین | جون | یكه |
| نهی | بهی | بهی | نی | بای | فين | اده | بان | اهه | ـبدا | اسی | ـبكه | ـبدی | دین | دیه | کیب | دبه | اوب | انب |
| ـیبا | نهی | هیل | کهب | بان | ـبقی | ـبان | انی | اهه | ـبده | بهی | ـبكه | دین | ـبهی | بهی | چین | کیب | اوب | ـكهی |
| ـبنا | ـهوه | یبل | بهج | ـنب | ئای | ئیس | ایس | یته | ـبده | هب | ادا | وچی | ـبهی | ـبیه | ـیبی | آبی | آنی | انی |
| بنا | ـوب | کرد | انی | ئنی | یمی | ـوست | بله | ـبه | ـوبن | أك | ـبنی | ـتبی | تهی | دین | بهی | دبی | ـبهن | ـكهی |
| داب | ـیهی | ـیه | نان | بری | ـبهن | ـبدا | ـیه | می... | ـبنی | كه | ـوچ | بهی | دین | بهی | ـ یكی | ـبجو | داب | ـبنا |

Table A.3: The 10 most frequent trigrams in the collected corpora of the selected languages. ے and _ represent space and ZWNJ, respectively. Among the trigrams, many affixes and conjunctions can be seen, such as ـوین ('and') in Northern Kurdish (KMR), ـكه ('that') in Gorani (HAC).

| Language | Noise % | Prediction (@1) | | Sentence |
|---|---|---|---|---|
| | | `lid.176` | Our's | |
| Punjabi | 0 | Urdu | Azeri | اور لادینیت واشتراكیت کو جمہوریت کے حسین لباہ میں پیش کردیا گیا ۔ |
| Saraiki | 0 | Punjabi | Saraiki | کیں وی زبان وادب تے تحقیق زیادہ تر کیفیتی |
| Sindhi | 0 | Sindhi | Sindhi | گهٹا دفعا هكـ عورت سائيٽي جنهن سان كوئي افلاطوني |
| Balochi | 0 | Urdu | Balochi | آیانی رابا که تئی مهر بوتگ آنت گنڇ گُوار |
| Azeri | 0 | Persian | Azeri | قوژی و دوغو سوریه موختار ایدارائتمہسی |
| Gilaki | 0 | Persian | Gilaki | شوراب ایسم ایته روستا ایسه جه راستوپی دهستان |
| Persian | 0 | Persian | Persian | جوانی زمان فرا گرفتن دانایی است. پیری زمان تمرین کردن آن است. |
| Uyghur | 0 | Uyghur | Uyghur | ھەيدەكچىلىك تەرتىپىنى ئاياغلاشتۇرۇشى توغرىسسدا كبسسم چىقىرىدۇ |
| Southern Kurdish | 0 | Sorani | Southern Kurdish | ڤایرۆس كۆرۆنا لەرێ دادوەر و پاریزەرەہیل دەوام لە دادگای ھەولێر وسان |
| Kashmiri | 20 | Urdu | Kashmiri | سودہا رانی چھ آكھ بندوستأنے أداكارہ یوس فِلمَن مَنۡز چھ كأم كَران. |
| Kashmiri | 100 | Urdu | Kashmiri | سودہا رانی چھ اكھ بندوستانی اداكارہ یوس فلمن منز چھ كام كران. |
| Sorani | 20 | Persian | Sorani | رێژی دەرجوانی ئهمسال لە سالی پێشتر زیاتەرہ |
| Sorani | 100 | Arabic | Sorani | رێژی دەرجانی ئهمسال لە صالی پیشطر زیاطرہ |

Table A.4: A few examples in the selected languages along with the predictions of fastText's pretrained models (`lid.176`) in comparison to those of one of our models trained using fastText on our collected data.